\documentclass{article}

\usepackage[final,nonatbib]{neurips_2019}

\usepackage[utf8]{inputenc} % allow utf-8 input
\usepackage[T1]{fontenc}    % use 8-bit T1 fonts
\usepackage{hyperref}       % hyperlinks
\usepackage{url}            % simple URL typesetting
\usepackage{booktabs}       % professional-quality tables
\usepackage{amsfonts}       % blackboard math symbols
\usepackage{nicefrac}       % compact symbols for 1/2, etc.
\usepackage{microtype}      % microtypography

\usepackage{amsmath}
\usepackage{amssymb}
\usepackage{mathtools}
\usepackage{amsfonts}
\usepackage{tabularx}
\usepackage{graphicx}
\usepackage{xfrac}
\usepackage{float}
\usepackage{wrapfig}
\usepackage{xcolor}
\usepackage{lipsum}
\usepackage{colortbl}
\usepackage{caption}

\newcommand{\cvec}{\mathbf{c}}
\newcommand{\svec}{\mathbf{s}}
\newcommand{\tvec}{\mathbf{t}}
\newcommand{\xvec}{\mathbf{x}}
\newcommand{\yvec}{\mathbf{y}}
\newcommand{\zvec}{\mathbf{z}}
\newcommand{\muvec}{\boldsymbol{\mu}}
\newcommand{\pivec}{\boldsymbol{\pi}}

\title{Integer Discrete Flows and Lossless Compression}

\author{%
    Emiel Hoogeboom\thanks{Equal contribution} \\
    UvA-Bosch Delta Lab  \\
    University of Amsterdam \\
    Netherlands \\
    \texttt{e.hoogeboom@uva.nl} \\
    \And
    Jorn W.T. Peters$^*$ \\
    UvA-Bosch Delta Lab \\
    University of Amsterdam \\
    Netherlands \\
    \texttt{j.w.t.peters@uva.nl} \\
    \AND
    Rianne van den Berg\thanks{Now at Google} \\
    University of Amsterdam \\
    Netherlands \\
    \texttt{riannevdberg@gmail.com} \\
    \And
    Max Welling \\
    UvA-Bosch Delta Lab \\
    University of Amsterdam \\
    Netherlands \\
    \texttt{m.welling@uva.nl} \\
}

\begin{document}

\maketitle

\begin{abstract}
Lossless compression methods shorten the expected representation size of data without loss of information, using a statistical model. Flow-based models are attractive in this setting because they admit exact likelihood optimization, which is equivalent to minimizing the expected number of bits per message. However, conventional flows assume continuous data, which may lead to reconstruction errors when quantized for compression. For that reason, we introduce a flow-based generative model for ordinal discrete data called Integer Discrete Flow (IDF): a bijective integer map that can learn rich transformations on high-dimensional data. As building blocks for IDFs, we introduce a flexible transformation layer called integer discrete coupling. Our experiments show that IDFs are competitive with other flow-based generative models. Furthermore, we demonstrate that IDF based compression achieves state-of-the-art lossless compression rates on CIFAR10, ImageNet32, and ImageNet64. To the best of our knowledge, this is the first lossless compression method that uses invertible neural networks. 
%   The abstract paragraph should be indented \nicefrac{1}{2}~inch (3~picas) on
%   both the left- and right-hand margins. Use 10~point type, with a vertical
%   spacing (leading) of 11~points.  The word \textbf{Abstract} must be centered,
%   bold, and in point size 12. Two line spaces precede the abstract. The abstract
%   must be limited to one paragraph.
% \lipsum[1]
\end{abstract}

\section{Introduction}
\label{sec:introduction}

Every day, 2500 petabytes of data are generated. Clearly, there is a need for compression to enable efficient transmission and storage of this data. Compression algorithms aim to decrease the size of representations by exploiting patterns and structure in data. In particular, \textit{lossless} compression methods preserve information perfectly--which is essential in domains such as medical imaging, astronomy, photography, text and archiving. 
Lossless compression and likelihood maximization are inherently connected through Shannon's source coding theorem~\cite{shannon1948mathematical}, i.e., the expected message length of an optimal entropy encoder is equal to the negative log-likelihood of the statistical model. In other words, maximizing the log-likelihood (of data) is equivalent to minimizing the expected number of bits required per message.

% Building a statistical model for compression can be challenging, because data is usually high-dimensional, which makes designing the likelihood and optimization difficult.

In practice, data is usually high-dimensional which introduces challenges when building statistical
models for compression. In other words, designing the likelihood and optimizing it for
high dimensional data is often difficult.
Deep generative models permit learning these complicated statistical models from data and have demonstrated their effectiveness in image, video, and audio modeling \cite{kingma2018glow,kumar2019videoflow, prenger2019waveglow}. Flow-based generative models \cite{dinh2014nice, dinh2016density,papamakarios2017masked,kingma2018glow,grathwohl2018ffjord,hoogeboom2019emerging} are advantageous over other generative models: \textit{i)} they admit exact log-likelihood optimization in contrast with Variational AutoEncoders (VAEs) \cite{kingma2014auto} and \textit{ii)} drawing samples (and decoding) is comparable to inference in terms of computational cost, as opposed to PixelCNNs \cite{van2016pixel}. However, flow-based models are generally defined for continuous probability distributions, disregarding the fact that digital media is stored discretely--for example, pixels from 8-bit images have 256 distinct values. In order to utilize continuous flow models for compression, the latent space must be quantized. This produces reconstruction errors in image space, and is therefore not suited for lossless compression.

\begin{figure}
    \centering
    \includegraphics[width=0.78\textwidth]{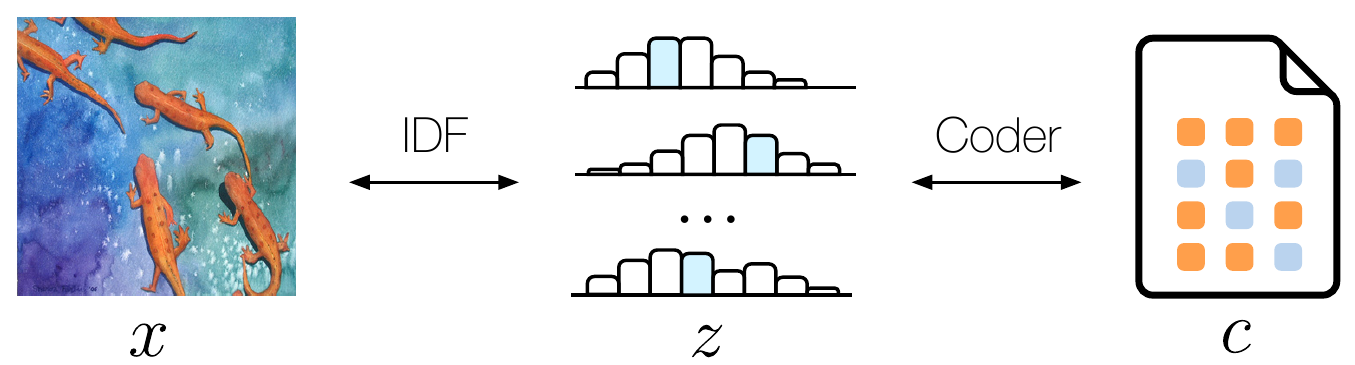}
    \caption{Overview of IDF based lossless compression. An image $x$ is transformed to a latent representation $z$ with a tractable distribution $p_Z(\cdot)$. An entropy encoder takes $z$ and $p_Z(\cdot)$ as input, and produces a bitstream $c$. To obtain $x$, the decoder uses $p_Z(\cdot)$ and $c$ to reconstruct $z$. Subsequently, $z$ is mapped to $x$ using the inverse of the IDF.}
    \label{fig:overview}
\end{figure}

To circumvent the (de)quantization issues, we propose Integer Discrete Flows (IDFs), which are invertible transformations for ordinal discrete data--such as images, video and audio. 
We demonstrate the effectiveness of IDFs by attaining state-of-the-art lossless compression performance on CIFAR10, ImageNet32 and ImageNet64. For a graphical illustration of the coding steps, see Figure~\ref{fig:overview}.
In addition, we show that IDFs achieve generative modelling results competitive with other flow-based methods. The main contributions of this paper are summarized as follows: \textit{1)} We introduce a generative flow for ordinal discrete data (Integer Discrete Flow), circumventing the problem of (de)quantization; \textit{2)} As building blocks for IDFs, we introduce a flexible transformation layer called integer discrete coupling; \textit{3)} We propose a neural network based compression method that leverages IDFs; and \textit{4)} We empirically show that our image compression method allows for progressive decoding that maintains the global structure of the encoded image. Code to reproduce the experiments is available at \url{https://github.com/jornpeters/integer_discrete_flows}.

\section{Background}
\label{sec:background}

The \textit{continuous} change of variables formula lies at the foundation of flow-based generative models. It admits exact optimization of a (data) distribution using a simple distribution and a learnable bijective map. Let $f: \mathcal{X} \to \mathcal{Z}$ be a bijective map, and $p_Z(\cdot)$ a prior distribution
on $\mathcal{Z}$. The model distribution $p_X(\cdot)$ can then be expressed as:
\begin{equation}
p_X(x) = p_Z(z) \left| \frac{dz}{dx} \right|,\quad \text{for \,$z = f(x)$}.
\label{eq:change_of_variables}  
\end{equation}
That is, for a given observation $x$, the likelihood is given by $p_Z(\cdot)$ evaluated at $f(x)$, normalized by the Jacobian determinant. A composition of invertible functions, which can be viewed as a repeated application of the change of variables formula, is generally referred to as a normalizing flow in the deep learning literature \cite{deco1995decorr, tabak2010density, tabak2013family, rezende2015norm}.

% Consider a bijective map between variables $x \in \mathbb{R}^d$ and $z \in \mathbb{R}^d$. The likelihood of $x$ can be expressed as the likelihood of $z = f(x)$ by evaluating $p_Z(z)$ and normalizing by the Jacobian determinant:
% \begin{equation}
% p_X(x) = p_Z(z) \left| \frac{dz}{dx} \right| \,\, ; \,\, z = f(x)
% \label{eq:change_of_variables}  
% \end{equation}

% A composition of invertible functions is generally referred to as a normalizing flow, which can be viewed as a repeated application of the change of variables formula \cite{deco1995decorr, tabak2010density, tabak2013family, rezende2015norm}. 

\subsection{Flow Layers}
The design of invertible transformations is integral to the construction of normalizing flows. In this section two important layers for flow-based generative modelling are discussed.

\textbf{Coupling layers} are tractable bijective mappings that are extremely flexible, when combined into a flow~\cite{dinh2016density,dinh2014nice}. Specifically, they have an analytical inverse, which is similar to a forward pass in terms of computational cost and the Jacobian determinant is easily computed, which makes coupling layers attractive for flow models. Given an input tensor $\xvec \in \mathbb{R}^d$, the input to a coupling layer is partitioned into two sets such that $\xvec = [\xvec_a,\ \xvec_b]$.
The transformation, denoted $f(\cdot)$, is then defined by:
\begin{align}
    \zvec = [\zvec_a,\ \zvec_b] = f(\xvec) = 
    [\xvec_a,\ \xvec_b \odot \svec(\xvec_a) + \tvec(\xvec_a)],
\end{align}
% They split an input $\xvec$ in two parts, $\xvec_a$ and $\xvec_b$. The output consists of a copy of the first part, $\zvec_a = \xvec_a$, and a transformation of the second part $\zvec_b = \xvec_b \odot s(\xvec_a) + t(\xvec_a)$
where $\odot$ denotes element-wise multiplication and $s$ and $t$ may be modelled using neural networks. Given this, the inverse is easily computed, i.e., $\xvec_a = \zvec_a$, and $\xvec_b = (\zvec_b - \tvec(\xvec_a))\oslash \svec(\xvec_a) $, where $\oslash$ denotes element-wise division. For $f(\cdot)$ to be invertible, $\svec(\xvec_a)$ must
not be zero, and is often constrained to have strictly positive values.

\textbf{Factor-out layers} allow for more efficient inference and hierarchical modelling. A general flow, following the change of variables formula, is described as a single map $\mathcal{X} \to \mathcal{Z}$. This implies that a $d$-dimensional vector is propagated throughout the whole flow model.
Alternatively, a part of the dimensions can already be \emph{factored-out} at regular intervals~\cite{dinh2016density},
such that the remainder of the flow network operates on lower dimensional data. We give an example for two levels ($L=2$) although this principle can be applied to an arbitrary number of levels:
\begin{gather}
\begin{aligned}
    [\zvec_1, \yvec_1] &= f_1(\xvec), \qquad \zvec_2 &= f_2(\yvec_1), \qquad \zvec &= [\zvec_1, \zvec_2],
\end{aligned}
\end{gather}
where $\xvec \in \mathbb{R}^d$ and $\yvec_1, \zvec_1, \zvec_2 \in \mathbb{R}^{d/2}$. 
The likelihood of $\xvec$ is then given by:
\begin{align}
    p(\xvec) &= 
    p(\zvec_2) \left| \frac{\partial f_2(\yvec_1)}{\partial\yvec_{1}} \right|
    p(\zvec_1|\yvec_1) \left| \frac{\partial f_1(\xvec)}{\partial\xvec} \right|.
\end{align}

% \begin{gather}
% \begin{aligned}
%     [\zvec_l, \yvec_l] &= f_l(\yvec_{l-1})\quad \text{for $l = 1,\ldots,L-1$,} \\
%     \zvec_L &= f_L(\yvec_{L-1}), \\
%     \zvec &= [\zvec_1, \ldots, \zvec_L],
% \end{aligned}
% \end{gather}
% where $\yvec_0 = \xvec \in \mathbb{R}^d$, $\yvec_1  \in \mathbb{R}^{d/2}$ for $l < L$, and $\yvec_L \in \mathbb{R}^{d/2^{L-1}}$, 
% The likelihood of $\xvec$ is then given by:
% \begin{align}
%     p(\xvec) &= 
%     p(\zvec_L) \left| \frac{\partial\zvec_L}{\partial\yvec_{L-1}} \right|
%     \prod_{l=1}^{L-1} p(\zvec_l|\yvec_l) \left| \frac{\partial f(\yvec_{l-1})}{\partial\yvec_{l-1}} \right|.
% \end{align}
This approach has two clear advantages. First, it admits a factored model for $\zvec$, $p(\zvec) = p(\zvec_L)p(\zvec_{L-1}|\zvec_L)\cdots p(\zvec_1|\zvec_2, \ldots, \zvec_L)$, which allows for conditional
dependence between parts of $\zvec$. This holds because the flow defines a bijective map between $\yvec_l$ and
$[\zvec_{l+1},\ldots, \zvec_L]$. Second, the lower dimensional flows are computationally more efficient.

% directly model a subset of the representation under a conditional prior \cite{dinh2014nice}. The remaining set is modelled by another flow and has lower dimensionality. This is advantageous because the prior can be more expressive because it is conditioned on the remaining set, and the lower dimensional flow is cheaper computationally. Let $\xvec \in \mathbb{R}^d$, and $[\zvec_1, \zvec_2] = \zvec \in \mathbb{R}^d$ and assume there exists some intermediate representation $[\yvec_1, \zvec_1] = f_1(\xvec)$ where $\yvec_1 \in \mathbb{R}^{<d}$. The representation $\yvec_1$ is modelled using another flow $\zvec_2 = f_2(\yvec_1)$ and $\zvec_1$ may be modeled as a distribution, conditioned on $\yvec_1$:
% \begin{align}
% \begin{split}
%   p(\xvec) = p(\zvec_1 | \yvec_1) &p(\yvec_1) \left| \frac{d\yvec_1}{d\xvec} \right| \,\, ; \,\, [\zvec_1, \yvec_1] = f_1(\xvec), \\
%   &p(\yvec_1) = p(\zvec_2) \left| \frac{d\zvec_2}{d\yvec_1} \right| \,\, ; \,\,  \zvec_2 = f_2(\yvec_1).
% \end{split}
% \end{align}

\subsection{Entropy Encoding}
Lossless compression algorithms map every input to a unique output and are designed to make \textit{probable} inputs \textit{shorter} and \textit{improbable} inputs \textit{longer}. Shannon's source coding theorem \cite{shannon1948mathematical} states that the optimal code length for a symbol $x$ is $-\log \mathcal{D}(x)$, and the minimum expected code length is lower-bounded by the entropy:
\begin{equation}
    \mathbb{E}_{x \sim \mathcal{D}} \left[|c(x)|\right] \approx \mathbb{E}_{x \sim \mathcal{D}} \left[ -\log p_X(x) \right] \geq \mathcal{H}(\mathcal{D}) ,
\end{equation}
where $c(x)$ denotes the encoded message, which is chosen such that $|c(x)| \approx -\log p_X(x)$, $|\cdot|$ is length, $\mathcal{H}$ denotes entropy, $\mathcal{D}$ is the data distribution, and $p_X(\cdot)$ is the statistical model that is used by the encoder. Therefore, maximizing the model log-likelihood is equivalent to minimizing the expected number of bits required per message, when the encoder is optimal.

Stream coders encode sequences of random variables with different probability distributions. They have near-optimal performance, and they can meet the entropy-based lower bound of Shannon \cite{rissanen1979arithmetic, moffat1998arithmetic}. 
In our experiments, the recently discovered and increasingly popular stream coder rANS~\cite{duda2013asymmetric} is used.
It has gained popularity due to its computational and coding efficiency. See Appendix~\ref{sec:asymmetric_numeral_systems}
for an introduction to rANS.

% In our experiments the entropy coder rANS~\cite{duda2013asymmetric} (see Appendix \ref{sec:asymmetric_numeral_systems}) is used, a recently discovered stream code which is becoming increasingly popular because of its computational and coding efficiency. 

% random variables with  with minimal overhead to terminate a message. approach the Shannon limit  as the length of a message increases:
% for any set of symbols and probability distributions, 

% \subsubsection{Likelihood and code length}
% In the case of complicated multivariate random variables (such as images), the true distribution $p(\xvec)$ may be unknown and is learned by a model $p_\theta(\xvec)$. The expected message length of an entropy encoder using $p_\theta$ as a model to encode messages $\xvec \sim \mathcal{D}$ is:
% \begin{equation}
%   \mathbb{E}_{\xvec \sim \mathcal{D}} \left[-\log p_{\theta}(\xvec)\right],
% \end{equation}

% which is identical to the negative log-likelihood objective. Therefore, a density model that is optimized for log-likelihood, is also optimized for the minimum expected code length.

\section{Integer Discrete Flows}
\label{sec:method}
We introduce Integer Discrete Flows (IDFs): a bijective integer map that can represent rich transformations. IDFs can be used to learn the probability mass function on (high-dimensional) ordinal discrete data. Consider an integer-valued observation $x \in \mathcal{X} = \mathbb{Z}^d$, a prior distribution $p_Z(\cdot)$ with support on $\mathbb{Z}^d$, and a bijective map $f: \mathbb{Z}^d \to \mathbb{Z}^d$ defined by an IDF. The model distribution $p_X(\cdot)$ can then be expressed as:
\begin{equation}
p_X(x) = p_Z(z),\quad z = f(x).
\label{eq:discrete_change_of_variables}  
\end{equation}
Note that in contrast to Equation \ref{eq:change_of_variables}, there is no need
for re-normalization using the Jacobian determinant. % and, contrary to standard flow-based transformations, there is no need to evaluate the Jacobian determinant.
Deep IDFs are obtained by stacking multiple IDF layers $\{f_l\}_{l=1}^L$, which are guaranteed to be bijective if the individual maps $f_l$ are all bijective. 
For an individual map to be bijective, it must be one-to-one and onto. Consider the bijective
map $f: \mathbb{Z} \to 2\mathbb{Z},\ x \mapsto 2x$. Although, this map is a bijection, it requires
us to keep track of the codomain of $f$, which is impracticable in the case of many dimensions and
multiple layers. Instead, we design layers to be bijective maps from $\mathbb{Z}^d$ to $\mathbb{Z}^d$, which ensures that the composition of layers and its inverse is closed on $\mathbb{Z}^d$.

\subsection{Integer Discrete Coupling}
\begin{wrapfigure}{r}{0.45\textwidth}
\centering
  \includegraphics[width=0.26\textwidth]{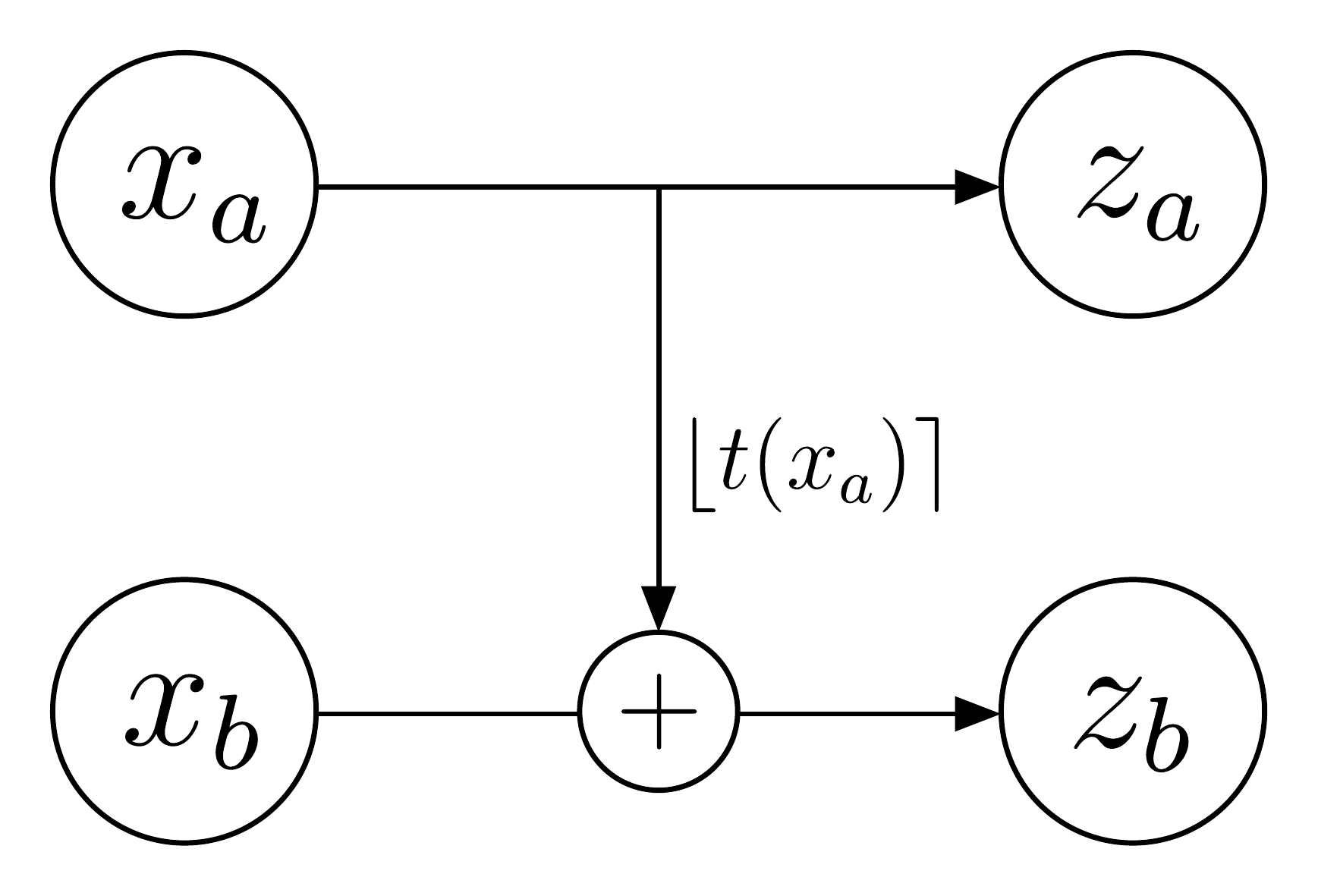}
    \caption{Forward computation of an integer discrete coupling layer. The input is split in two parts. The output consists of a copy of the first part, and a conditional transformation of the second part. The inverse of the coupling layer is computed by inverting the conditional transformation.}
    \label{fig:discrete_coupling}
\end{wrapfigure}

As a building block for IDFs, we introduce integer discrete coupling layers. These are invertible and the set $\mathbb{Z}^d$ is closed under their transformations. Let $[\xvec_a, \xvec_b] = \xvec \in \mathbb{Z}^d$ be an input of the layer.
The output $\zvec = [\zvec_a, \zvec_b]$ is defined as a copy $\zvec_a = \xvec_a$, and a transformation $\zvec_b = \xvec_b + \lfloor \tvec(\xvec_a) \rceil$, where $\lfloor \cdot \rceil$ denotes a nearest rounding operation and $\tvec$ is a neural network (Figure \ref{fig:discrete_coupling}).

Notice the multiplication operation in standard coupling is not used in integer discrete coupling, because it does not meet our requirement that the image of the transformations is equal to $\mathbb{Z}$. It may seem disadvantageous that our model only uses translation, also known as additive coupling, however, large-scale continuous flow models in the literature tend to use \textit{additive} coupling instead of affine coupling \cite{kingma2018glow}. 

In contrast to existing coupling layers, the input is split in 75\%--25\% parts for $\xvec_a$ and $\xvec_b$, respectively. As a consequence, rounding is applied to fewer dimensions, which results in less gradient bias. In addition, the transformation is richer, because it is conditioned on more dimensions. Empirically this results in better performance. 

\textbf{Backpropagation through Rounding Operation}
As shown in Figure~\ref{fig:discrete_coupling}, a coupling layer in IDF requires a rounding operation ($\lfloor\cdot\rceil$) on the predicted translation. Since the rounding operation is effectively a step function, its
gradient is zero almost everywhere. As a consequence, the rounding operation is inherently
incompatible with gradient based learning methods. In order to backpropagate through the rounding operations, we make use of the Straight Through Estimator (STE)~\cite{bengio2013estimating}.
% This issue if often faced in the
% binary and/or quantized Neural Networks literature. Roughly, one can identify two approaches
% to handle this: (1) use a (relaxed) stochastic objective [\textcolor{red}{citation}] or (2) leverage the straight-through estimator (STE)~\cite{bengio2013estimating}.
% Experimentally, we found that STE-based models showed robust training behaviour and resulted in the best performing models. \textcolor{red}{make shorter}
In short, the STE \emph{ignores} the rounding operation during
back-propagation, which is equivalent to redefining the gradient of the rounding operation as follows:
\begin{equation}
\nabla_{\boldsymbol{x}} \lfloor \boldsymbol{x} \rceil \triangleq \boldsymbol{I}.
%, \quad
%\nabla_{\boldsymbol{x}} g(\lfloor \boldsymbol{x} \rceil) \triangleq \nabla_{\boldsymbol{x}} g(\boldsymbol{x})
\end{equation}

\textbf{Lower Triangular Coupling}\\
There exists a trade-off between the number of integer discrete coupling layers and the complexity of the layers in IDF architectures, due to the gradient bias that is introduced by the rounding operation (see section \ref{sec:tradeoff}). We introduce a \textit{multivariate} coupling transformation called Lower Triangular Coupling, which is specifically designed such that the number of rounding operations remains unchanged. For more details, see Appendix~\ref{appendix:lowerl}.

\begin{wrapfigure}[10]{r}{0.35\textwidth}
    \vspace{-8mm}
    \centering
    \includegraphics[width=0.3\textwidth]{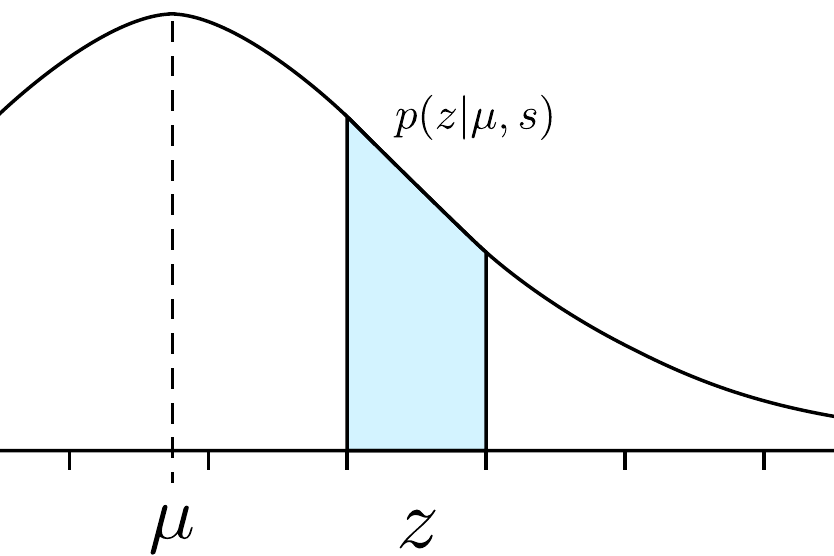}
    \caption{The discretized logistic distribution. The shaded area shows the probability density.}
    \label{fig:discrete distribution}
\end{wrapfigure}
\subsection{Tractable Discrete distribution}
As discussed in Section~\ref{sec:background}, a simple distribution $p_Z(\cdot)$ is posed on $\mathcal{Z}$ in flow-based models. In IDFs, the prior $p_Z(\cdot)$ is a factored discretized logistic distribution (DLogistic) \cite{kingma2016improved, salimans2017pixelcnn++}. The discretized logistic captures the inductive bias that values close together are related, which is well-suited for ordinal data. %This also avoids issues in extreme cases, where agnostic models will place zero probability on $1$ if only values $0$ and $2$ are observed 

%An important difference with literature is that traditionally only mean $\mu$ and scale $s$ are optimized, but we also require gradients with respect to the variable $z$. 

The probability mass $\text{DLogistic}(z|\mu, s)$ for an integer $z \in \mathbb{Z}$,
mean $\mu$, and scale $s$ is defined as the density assigned to the interval $[z-\frac12, z+\frac12]$ by the probability density function of $\text{Logistic}(\mu, s)$ (see Figure \ref{fig:discrete distribution}). This can be efficiently computed by evaluating the cumulative distribution function twice:
% The probability density of an integer valued $z \in \mathbb{Z}$ is defined as an integral of the probability distribution over the bin (see Figure \ref{fig:discrete distribution}), which can be efficiently obtained by evaluating the cumulative distribution function twice:
\begin{align}
\begin{split}
    % p_Z(Z=z| \mu, s) &= 
    \text{DLogistic}(z|\mu, s) = 
    \int_{z-\frac{1}{2}}^{z+\frac{1}{2}} \text{Logistic}(z' | \mu, s) \mathrm{d}z'
    % &= \left. \sigma\left(\frac{z' - \mu}{s}\right) \right\rvert_{z-\frac{1}{2}}^{z+\frac{1}{2}},
    = \sigma\left(\frac{z + \frac12 - \mu}{s}\right) - \sigma\left(\frac{z - \frac12 - \mu}{s}\right),
\end{split}
\end{align}

where $\sigma(\cdot)$ denotes the sigmoid, the cumulative distribution function of a standard Logistic.
In the context of a factor-out layer, the mean $\mu$ and scale $s$ are conditioned on the subset of
\begin{wrapfigure}[45]{r}{0.4\textwidth}
    \centering
    \vspace{0mm}
    \includegraphics[width=0.4\textwidth]{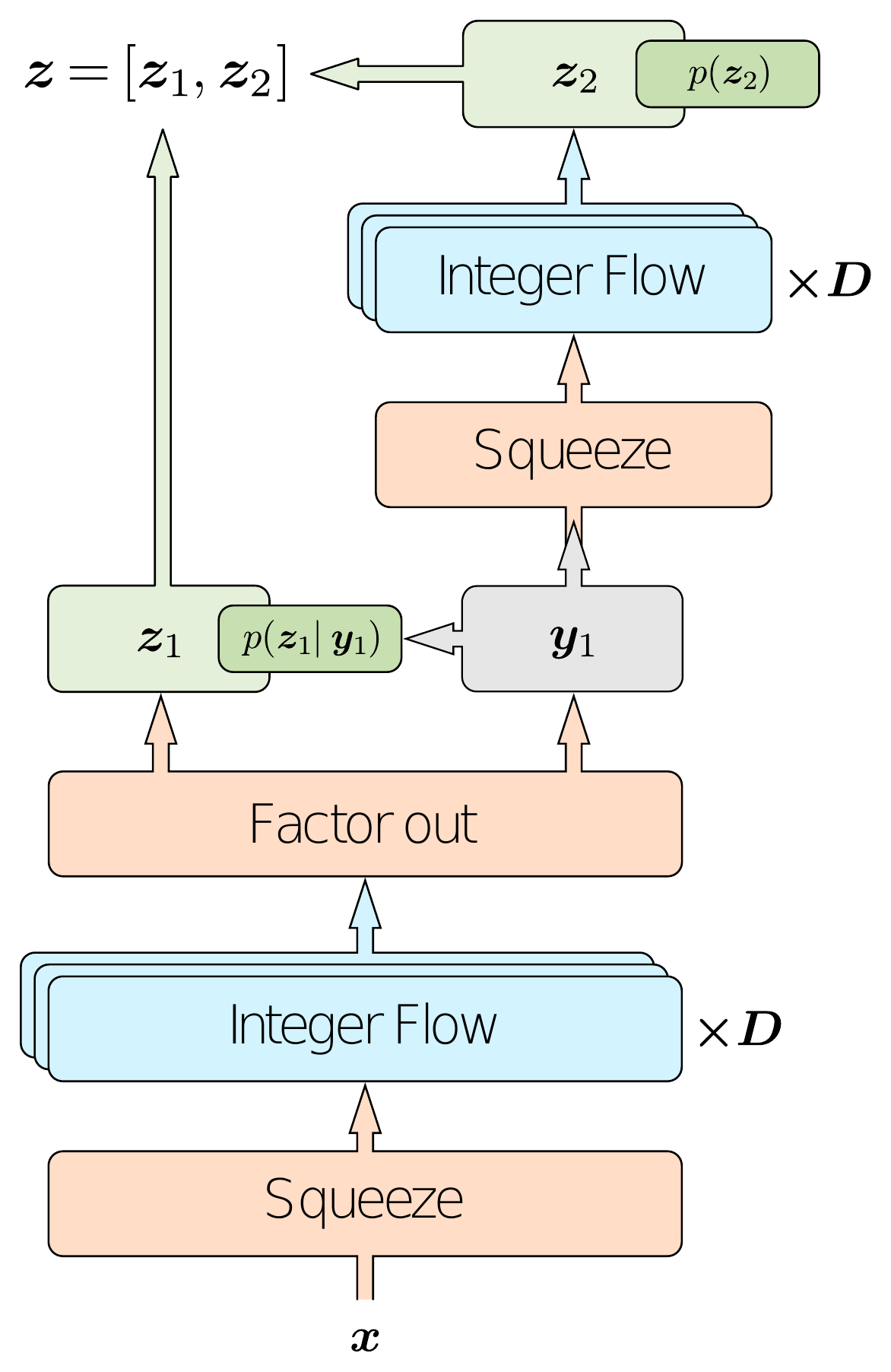}
    \caption{Example of a 2-level flow architecture. The squeeze layer reduces the spatial dimensions by two, and increases the number of channels by four. A single integer flow layer consists of a channel permutation and an integer discrete coupling layer. Each level consists of $D$ flow layers.}
    \label{fig:model_layout}
% \end{wrapfigure}

% \begin{wrapfigure}[16]{r}{0.4\textwidth}
    \centering
    \vspace{5mm}
    \includegraphics[width=0.4\textwidth]{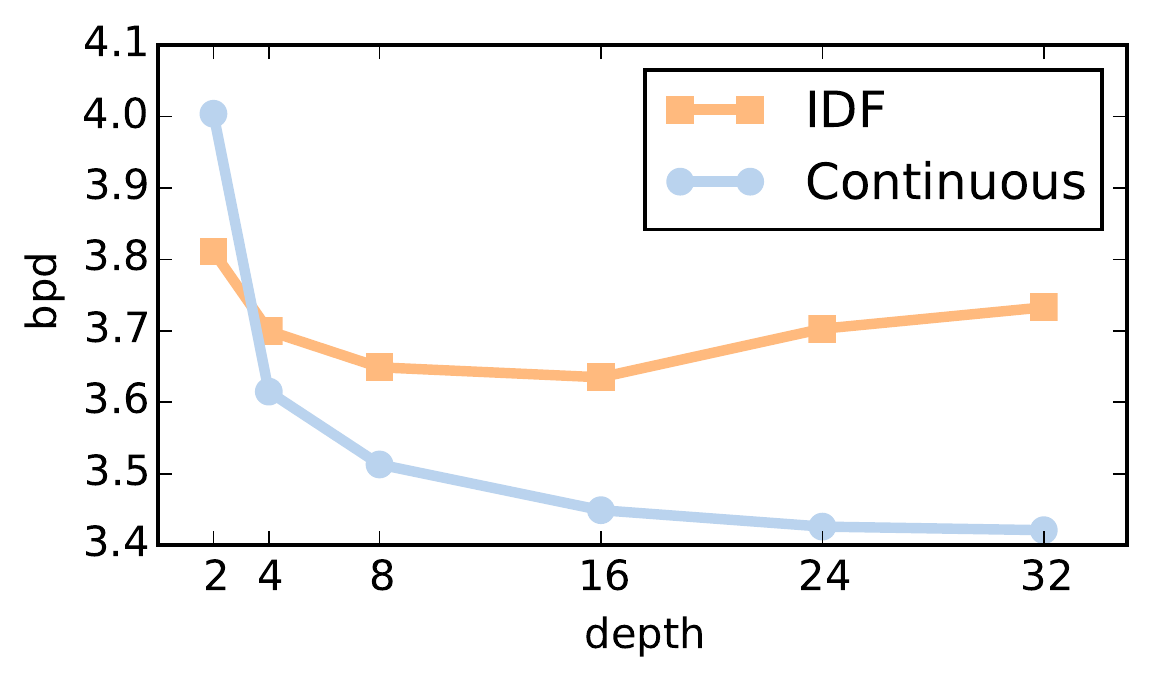}
    \caption{Performance of flow models for different depths (i.e. coupling layers per level). The networks in the coupling layers contain 3 convolution layers. Although performance increases with depth for continuous flows, this is not the case for discrete flows.}
    \label{fig:depth_performance}
\end{wrapfigure}
data that is \emph{not} factored out. That is, the input to the $l$th factor-out layer is split into 
$\zvec_l$ and $\yvec_l$. The conditional distribution on $\zvec_{l,i}$ is then given as $\text{DLogistic}(\zvec_{l,i}|\boldsymbol{\mu}(\yvec_l)_i, \textbf{s}(\yvec_l)_i)$, where $\boldsymbol{\mu}(\cdot)$ and $\textbf{s}(\cdot)$ are parametrized as neural networks.

\textbf{Discrete Mixture distributions}
The discretized logistic distribution is unimodal and therefore limited in complexity. With a marginal increase in computational cost, we increase the flexibility of the latent prior on $\zvec_L$ by extending it to a mixture of $K$ logistic distributions \cite{salimans2017pixelcnn++}:  
\begin{equation}
    p(z | \muvec, \svec, \pivec) = \sum_k^K \pi_{k} \cdot p(z | \mu_{k}, s_{k}).
\end{equation}
Note that as $K \to \infty$, the mixture distribution can model arbitrary univariate discrete distributions. In practice, we find that a limited number of mixtures ($K = 5$) is usually sufficient for image density modelling tasks.

\subsection{Lossless Source Compression}
Lossless compression is an essential technique to limit the size of representations without destroying information. Methods for lossless compression require \textit{i)} a statistical model of the source, and \textit{ii)} a mapping from source symbols to bit streams.

IDFs are a natural statistical model for lossless compression of ordinal discrete data, such as images, video and audio. They are capable of modelling complicated high-dimensional distributions, and they provide error-free reconstructions when inverting latent representations. The mapping between symbols and bit streams may be provided by any entropy encoder. Specifically, stream coders can get arbitrarily close to the entropy regardless of the symbol distributions, because they encode entire sequences instead of a single symbol at a time. 

In the case of compression using an IDF, the mapping $f: \xvec \mapsto \zvec$ is defined by the IDF.
Subsequently, $\zvec$ is encoded under the distribution $p_Z(\zvec)$ to a bitstream $\cvec$ using an entropy encoder. Note that, when using factor-out layers, $p_Z(\zvec)$ is also defined using the IDF.
Finally, in order to decode a bitstream $\cvec$, an entropy encoder uses $p_Z(\zvec)$ to obtain $\zvec$. and the original image is obtained by using the map $f^{-1}: \zvec \mapsto \xvec$, i.e., the inverse IDF. See Figure~\ref{fig:overview} for a graphical depiction of this process.

% Specifically, encoding an input image $\xvec$ requires a mapping provided by an IDF, $\zvec = f(\xvec)$, which is consequently encoded under the distribution $p(\zvec)$ to a bitstream $\cvec$ using an entropy encoder. To decode a bitstream $\cvec$, an entropy encoder uses $p(\zvec)$ to obtain $\zvec$. The original image is obtained by computing $\xvec = f^{-1}(\zvec)$.

In rare cases, the compressed file may be larger than the original. Therefore, following established practice in compression algorithms, we utilize an \emph{escape bit}. That is, the encoder will decide whether to encode the message or save it in raw format and encode that decision into the first bit. 

\section{Architecture}
\label{sec:architecture}
The IDF architecture is split up into one or more levels. Each level consists of a squeeze operation~\cite{dinh2016density},
$D$ integer flow layers, and a factor-out layer. Hence, each level defines a mapping
from $\yvec_{l-1}$ to $[\zvec_l, \yvec_l]$, except for the final level $L$, which defines a mapping
$\yvec_{L-1} \mapsto \zvec_L$. Each of the $D$ integer flow layers per level consist of a permutation layer followed by an integer discrete coupling layer.
Following~\cite{dinh2016density}, the permutation
layers are initialized once and kept fixed throughout training and evaluation. Figure~\ref{fig:model_layout} shows a graphical illustration of a two level IDF. The specific architecture details for each experiment are presented in Appendix~\ref{sec:networks}. In the remainder of this section, we discuss the trade-off between network depth and performance when rounding operations are used.

\subsection{Flow Depth and Network Depth}
\label{sec:tradeoff}
The performance of IDFs depends on a trade-off between complexity and gradient bias, influenced by the number of rounding functions. Increasing the performance of standard normalizing flows is often achieved by increasing the depth, i.e., the number of flow-modules. However, for IDFs each flow-module results in additional rounding operations that introduce gradient bias. As a consequence, 
adding more flow layers hurts performance, after some point, as is depicted in Figure \ref{fig:depth_performance}. We found that the limitation of using fewer coupling layers
in an IDF can be negated by increasing the complexity of the neural networks part of the coupling
and factor-out layers. That is, we use DenseNets~\cite{huang2017densely} in order to
predict the translation $\mathbf{t}$ in the integer discrete coupling layers
and $\mu$ and $s$ in the factor-out layers.
% performance of similar flow-based methods
% can be obtained by increasing the complexity of the neural network part of the coupling and
% factor-out layers.

% at some point adding more layers actually hurts performances, as is depicted in Figure \ref{fig:depth_performance}. Instead, it is more advantageous to increase the complexity of the network \textit{within} the coupling layers.

\section{Related Work}
\label{sec:related_work}

There exist several deep generative modelling frameworks. This work builds mainly upon flow-based generative models, described in \cite{rippel2013high, dinh2014nice, dinh2016density}. In these works, invertible functions for continuous random variables are developed. However, quantizing a latent representation, and subsequently inverting back to image space may lead to reconstruction errors \cite{dewitte1997lossless, calderbank1997lossless, calderbank1998wavelet}.

Other likelihood-based models such as PixelCNNs \cite{van2016pixel} utilize a decomposition of conditional probability distributions. However, this decomposition assumes an order on pixels which may not reflect the actual generative process. Furthermore, drawing samples (and decoding) is generally computationally expensive. VAEs \cite{kingma2014auto} optimize a lower bound on the log likelihood instead of the exact likelihood. They are used for lossless compression with deterministic encoders \cite{mentzer2019practical} and through bits-back coding. However, the performance of this approach is bounded by the lower bound. Moreover, in bits back coding a single data example can be inefficient to compress, and the \emph{extra bits} should be random, which is not the case in practice and may also lead to coding inefficiencies~\cite{townsend2019practical}.

Non-likelihood based generative models tend to utilize Generative Adversarial Networks \cite{goodfellow2014generative}, and can generate high-quality images. However, since GANs do not optimize for likelihood, which is directly connected to the expected number of bits in a message, they are not suited for lossless compression.

In the lossless compression literature, numerous reversible integer to integer transforms have been proposed \cite{ahmed1974discrete, dewitte1997lossless, calderbank1997lossless, calderbank1998wavelet}. Specifically, lossless JPEG2000 uses a reversible integer wavelet transform \cite{jp2_standard}. However, because these transformations are largely hand-designed, they are difficult to tune for real-world data, which may require complicated nonlinear transformations. 
% Lossless compression: Integer Wavelet transform in jp2. rANS encoder. 

Around time of submission, unpublished concurrent work appeared~\cite{discrete2019tran} that explores discrete flows. The main differences between our method and this work are: \textit{i)} we propose discrete flows for ordinal discrete data (e.g. audio, video, images), whereas they are are focused on categorical data. \textit{ii)} we provide a connection with the source coding theorem, and present a compression algorithm. \textit{iii)} We present results on more large-scale image datasets.
% Recently a paper with similar work 

% Differences:
% -Proposed for categorical discrete data.
% -Our work shows competitive performance on image data.
% -

\section{Experiments}
\label{sec:experiments}
To test the compression performance of IDFs, we compare with a number of established lossless compression methods: PNG~\cite{png_standard};  JPEG2000~\cite{jp2_standard}; FLIF~\cite{sneyers2016flif}, a recent format that uses machine learning to build decision trees for efficient coding; and Bit-Swap~\cite{kingma2019bit}, a VAE based lossless compression method. We show that IDFs outperform all these formats on CIFAR10, ImageNet32 and ImageNet64. In addition, we demonstrate that IDFs can be very easily tuned for specific domains, by compressing the ER + BCa histology dataset. For the exact treatment of datasets and optimization procedures, see Section~\ref{appendix:experimental_details}.

%Moreover, we explore \textcolor{red}{something on partial decoding}

\subsection{Image Compression}
The compression performance of IDFs is compared with competing methods on standard datasets, in bits per dimension and compression rate. The IDFs and Bit-Swap are trained on the train data, and compression performance of all methods is reported on the test data in Table~\ref{tab:compression_performance_datasets}. IDFs achieve state-of-the-art lossless compression performance on all datasets.

Even though one can argue that a compressor should be tuned for the source domain, the performance of IDFs is also examined on out-of-dataset examples, in order to evaluate compression generalization. We utilize the IDF trained on Imagenet32, and compress the CIFAR10 and ImageNet64 data. For the latter, a single image is split into four $32\times 32$ patches.  Surprisingly, the IDF trained on ImageNet32 (IDF$^\dagger$) still outperforms the competing methods showing only a slight decrease in compression performance on CIFAR10 and ImageNet64, compared to its source-trained counterpart.

As an alternative method for lossless compression, one could quantize the distribution $p_Z(\cdot)$ and the latent space $\mathcal{Z}$ of a continuous flow. This
results in reconstruction errors that need to be stored in addition to the latent representation $\zvec$, such that the original data can be recovered perfectly. We show that this scheme is ineffective for lossless compression. Results are presented in Appendix~\ref{appendix:quantizing_nf}.

% \newcolumntype{a}{>{\columncolor{gray!15!white}}l}
% \setlength{\aboverulesep}{0pt}
% \setlength{\belowrulesep}{0pt}
% \setlength{\extrarowheight}{.75ex}
\begin{table}
    \centering
    \caption{Compression performance of IDFs on CIFAR10, ImageNet32 and ImageNet64 in bits per dimension, and compression rate (shown in parentheses). The Bit-Swap results are retrieved from \cite{kingma2019bit}. The column marked IDF$^\dagger$ denotes an IDF trained on ImageNet32 and evaluated on the other datasets.}
    \scalebox{0.87}{
    \begin{tabular}{l l l l l l l}
    \toprule
        Dataset & IDF & IDF$^\dagger$ & Bit-Swap & FLIF~\cite{sneyers2016flif} & PNG & JPEG2000 \\ \midrule 
        CIFAR10 & \textbf{3.34 (2.40$\times$)} &
            3.60 (2.22$\times$) &
            3.82 (2.09$\times$) &
            4.37 (1.83$\times$) &
            5.89 (1.36$\times$) &
            5.20 (1.54$\times$)
            % 6.94 (1.15$\times$)
            \\ 
        ImageNet32 & \textbf{4.18 (1.91$\times$)} & 
        \textbf{4.18 (1.91$\times$)} & %
            4.50 (1.78$\times$) &
            5.09 (1.57$\times$) &
            6.42 (1.25$\times$) &
            6.48 (1.23$\times$) 
            % 6.98 (1.15$\times$)
            \\
        ImageNet64 & \textbf{3.90 (2.05$\times$)} & 
        3.94 (2.03 $\times$) &
            \multicolumn{1}{c}{--} &
            4.55 (1.76$\times$) &
            5.74 (1.39$\times$) &
            5.10 (1.56$\times$)
            % 6.72 (1.19$\times$)
            \\ 
        \bottomrule
    \end{tabular}}
    \label{tab:compression_performance_datasets}
\end{table}

\subsection{Tuneable Compression}
\begin{figure}
\centering
\begin{minipage}[t]{.61\textwidth}
  \centering
  \includegraphics[width=0.48\textwidth]{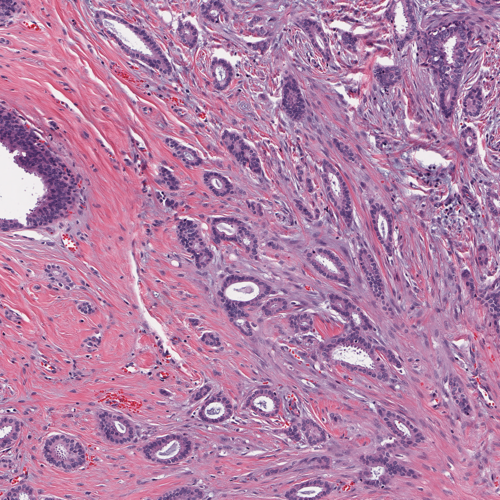}
    \hspace{0.01\textwidth}
    \includegraphics[width=0.48\textwidth]{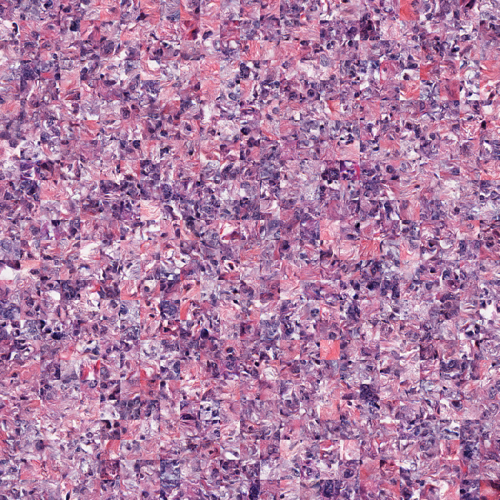}
  \captionof{figure}{Left: An example from the ER + BCa histology dataset. Right: 625 IDF samples of size 80$\times$80px.}
  \label{fig:dataset_histology}
\end{minipage}%
\hspace{0.01\textwidth}
\begin{minipage}[t]{.3\textwidth}
  \centering
  \includegraphics[width=0.975\textwidth]{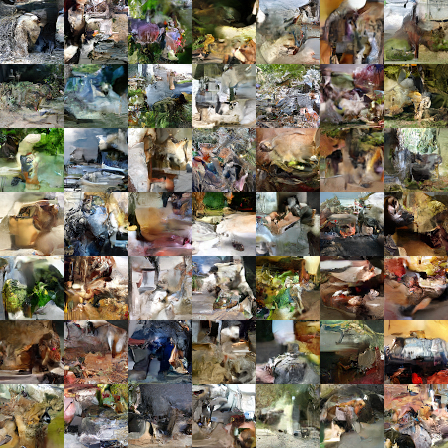}
  \captionof{figure}{49 samples from the ImageNet 64$\times$64 IDF.}
  \label{fig:samples_imagenet32}
\end{minipage}
\end{figure}

Thus far, IDFs have been tested on standard machine learning datasets. In this section, IDFs are tested on a specific domain, medical images. In particular, the ER + BCa histology dataset~\cite{janowczyk2018resolution} is used, which contains 141 regions of interest scanned at $40\times$, where each image is $2000 \times 2000$ pixels (see Figure~\ref{fig:dataset_histology}, left). Since current hardware does not support training on such large images directly, the model is trained on random $80\times80$px patches. See Figure \ref{fig:dataset_histology}, right for samples from the model. Likewise, the compression is performed in a patch-based manner, i.e., each patch is compressed independently of all other patches. IDFs are again compared with FLIF and JPEG2000, and also with a modified version of JPEG2000 that has been optimized for virtual microscopy specifically, named JP2-WSI~\cite{helin2018optimized}. Although the IDF is at a disadvantage because it has to compress in patches, it considerably outperforms the established formats, as presented in Table~\ref{tab:histology_performance}. 

\begin{table}
    \centering
    \caption{Compression performance on the ER + BCa histology dataset in bits per dimension and compression rate. JP2-WSI is a specialized format optimized for virtual microscopy.}
    \begin{tabular}{l l l l l}
    \toprule
        Dataset & IDF & JP2-WSI & FLIF~\cite{sneyers2016flif} & JPEG2000 \\ 
    \midrule
        Histology & \textbf{2.42 (3.19$\times$)} & 3.04 (2.63$\times$) & 4.00 (2.00$\times$) & 4.26 (1.88$\times$) \\ 
    \bottomrule
    \end{tabular}
    \label{tab:histology_performance}
\end{table}

\subsection{Progressive Image Rendering}
\begin{figure}
% \vspace{-15pt}
    \centering
    \includegraphics[width=0.92\textwidth]{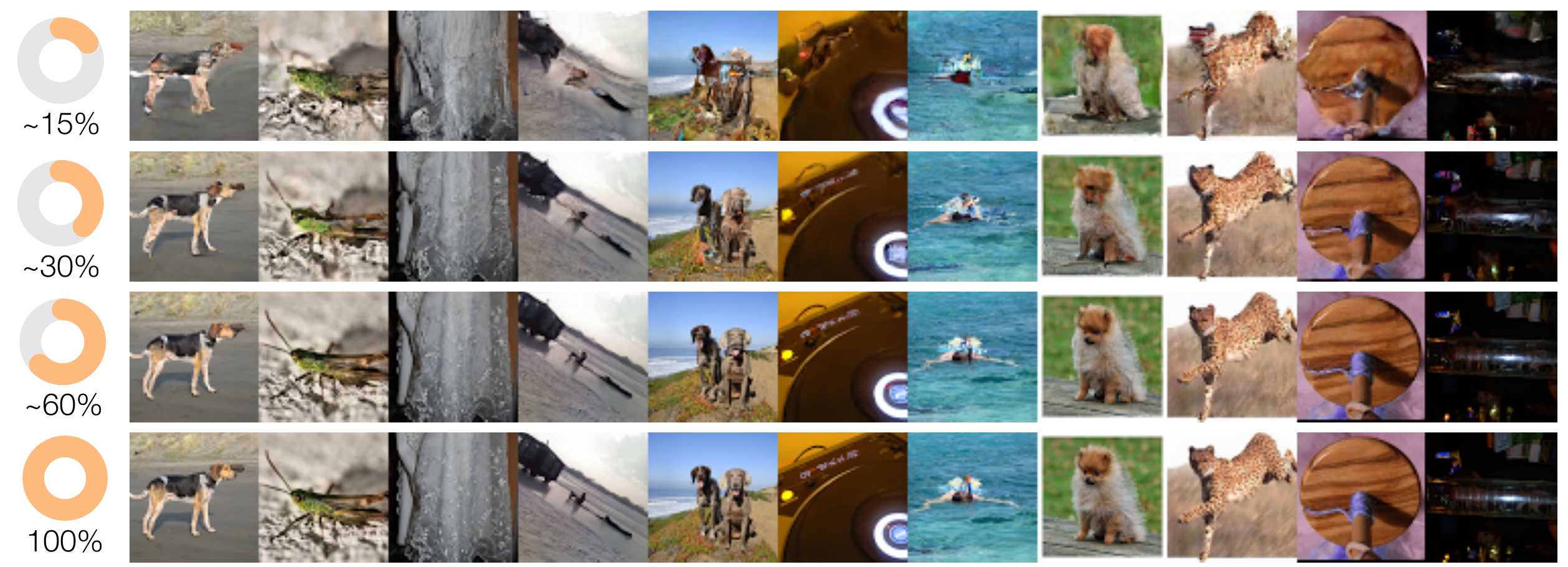}
    \vspace{-5pt}
    \caption{Progressive display of the data stream for images taken from the test set of ImageNet64. From top to bottom row, each image uses approximately 15\%, 30\%, 60\% and 100\% of the stream, where the remaining dimensions are sampled. Best viewed electronically.}
    \label{fig:progressive}
\end{figure}
In general, transferring data may take time because of slow internet connections or disk I/O. For this reason, it is desired to progressively visualize data, i.e., to render the image with more detail as more data arrives. Several graphics formats support progressive loading. However, the encoded file size may increase by enabling this option, depending on the format \cite{png_standard}, whereas IDFs support progressive rendering naturally. To partially render an image using IDFs, first the received variables are decoded. Next, using the hierarchical structure of the prior and ancestral sampling, the remaining dimensions are obtained. The progressive display of IDFs for ImageNet64 is presented in Figure~\ref{fig:progressive}, where the rows use approximately 15\%, 30\%, 60\%, and 100\% of the bitstream. The global structure is already captured by smaller fragments of the bitstream, even for fragments that contain only 15\% of the stream.  

% % \begin{figure}
% \begin{wrapfigure}{r}{0.38\textwidth}
% % \vspace{-15pt}
%     \centering
%     \includegraphics[width=0.35\textwidth]{images/partial_load.pdf}
%     \vspace{-10pt}
%     \caption{Progressive display of data stream. The first image uses approximately 30\%, the second 60\% and the last 100\% of the data.}
%     \label{fig:progressive}
% \end{wrapfigure}
% % \end{figure}

\subsection{Probability Mass Estimation}

In addition to a statistical model for compression, IDFs can also be used for image generation and probability mass estimation. Samples are drawn from an ImageNet 32$\times$32 IDF and presented in Figure \ref{fig:samples_imagenet32}. IDFs are compared with recent flow-based generative models, RealNVP \cite{dinh2016density}, Glow \cite{kingma2018glow}, and Flow++ in analytical bits per dimension (negative log$_2$-likelihood). To compare architectural changes, we modify the IDFs to \textit{Continuous} models by dequantizing, disabling rounding, and using a continuous prior. The continuous versions of IDFs tend to perform slightly better, which may be caused by the gradient bias on the rounding operation. IDFs show competitive performance on CIFAR10, ImageNet32, and ImageNet64, as presented in Table \ref{tab:generative_performance}. Note that in contrast with IDFs, RealNVP uses scale transformations, Glow has $1 \times 1$ convolutions and actnorm layers for stability, and Flow++ uses the aforementioned, and an additional flow for dequantization. Interestingly, IDFs have comparable performance even though the architecture is relatively simple.

% To study the effect of architectural changes, we modify the IDF to a `continuous' model by: changing the prior from discrete to a continuous distribution, removing rounding in coupling layers and, adding dequantization noise. 

\begin{table}
    \centering
    \vspace{-3mm}
    \caption{Generative modeling performance of IDFs and comparable flow-based methods in bits per dimension (negative log$_2$-likelihood).}
    \begin{tabular}{l l l l l l}
    \toprule
        Dataset & IDF & Continuous & RealNVP & Glow & Flow++  \\
    \midrule
        CIFAR10 & 3.32 & 3.31 & 3.49 & 3.35 & 3.08  \\ 
        ImageNet32 & 4.15 & 4.13 & 4.28 & 4.09 & 3.86  \\
        ImageNet64 & 3.90 & 3.85 & 3.98 & 3.81 & 3.69  \\ 
    \bottomrule
    \end{tabular}
    \label{tab:generative_performance}
\end{table}

\section{Conclusion}
\label{sec:conclusion}
We have introduced Integer Discrete Flows, flows for ordinal discrete data that can be used for deep generative modelling and neural lossless compression. We show that IDFs are competitive with current flow-based models, and that we achieve state-of-the-art lossless compression performance on CIFAR10, ImageNet32 and ImageNet64. To the best of our knowledge, this is the first lossless compression method that uses invertible neural networks.

\bibliographystyle{plain}
\bibliography{main.bib} 

\newpage
\appendix
\section{Additional background}

% \subsection{Lifting scheme}
% A lifting scheme \cite{sweldens1996lifting} is a technique that may be used to construct invertible wavelets. The transformation consists of a Predict (P) and update (U) step, and is invertible by reversing computation and inverting signs (see Figure \ref{fig:lifting}). Notably, the structure of the lifting scheme is very similar to that of a coupling layer. A noteworthy difference is that coupling layers do not include an update step.

% \begin{figure}[h]
%     \centering
%     \includegraphics[width=0.24\textwidth]{images/lifting.pdf}
%     \caption{A lifting scheme}
%     \label{fig:lifting}
% \end{figure}

% Invertible wavelet transforms that are confined to integers have been successfully used to losslessy encode images \cite{calderbank1998wavelet}. 

\subsection{Asymmetric Numeral Systems}
\label{sec:asymmetric_numeral_systems}
Asymmetric Numeral Systems (ANS) \cite{duda2009asymmetric} is a recent approach to entropy coding. The range-based variant: rANS, is generally used as a faster replacement for arithmetic coding, because a state is only represented by a single number and fewer mathematical operations are required \cite{duda2013asymmetric}.

\begin{figure}[h]
    \centering
    \includegraphics[width=0.45\textwidth]{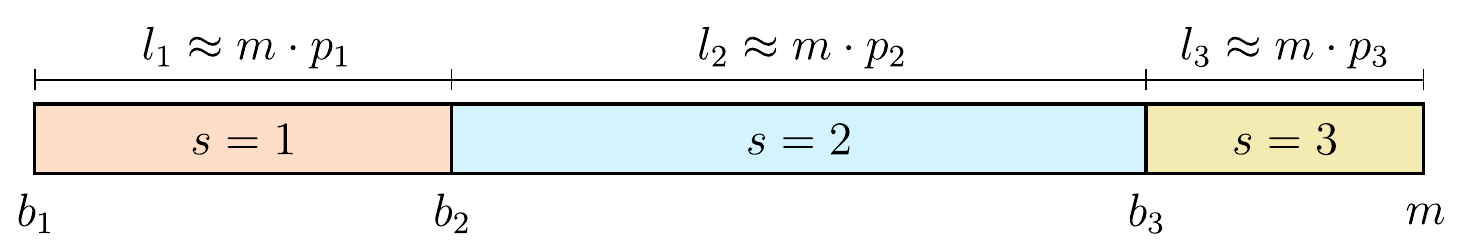}
    \caption{The unique sequences for each symbol}
    \label{fig:sequences}
\end{figure}

The encoding function of rANS encodes a symbol $s$ into a code $c'$ given the so far existing code $c$:
\begin{equation}
c'(c, s) = \lfloor c / l_s \rfloor \cdot m + (c \text{ mod } l_s) + b_s,
\end{equation}
where $m$ is a large integer that functions as the quantization denominator. Integers are chosen for $l_s$ such that $p(s) \approx \sfrac{l_s}{m}$, where $p(s)$ denotes the probability of symbol $s$. Each symbol is associated with a unique interval $[b_s, b_s + l_s)$, where $b_s = \sum_{i=1}^{s-1} l_i$, as depicted in Figure \ref{fig:sequences}.

The decoding function needs to retrieve the encoded symbol $s$, and the previous state $c$ from the new code $c'$. First consider the term $c' \text{ mod } m$, which is equal to the last two terms of the encoding function: $c \text{ mod } l_s + b_s$. This term is guaranteed to lie in the interval $[b_s$, $b_s + l_s)$. Therefore, the symbol can be retrieved by finding:
\begin{equation}
s(c') = t \,\, \text{  s.t. } b_{t} \leq c' \text{ mod } m < b_{t+1}.
\end{equation}

Consequently with the knowledge of $s$, the previous state $c$ can be obtained by computing:
\begin{equation}
c(c', s) = l_s \cdot \lfloor c' / m \rfloor + (c' \text{ mod } m) - b_s.
\end{equation}

In practice, $m$ is chosen as a power of two (for example $2^{32}$). As such, multiplication and division with $m$ reduces to bit shifts and modulo $m$ reduces to a binary masking operation.

\section{Lower Triangular Coupling} 
\label{appendix:lowerl}
There exists a trade-off between the number of integer discrete coupling layers and the complexity of the layers in IDF architectures, due to the gradient bias that is introduced by the rounding operation. For this reason, it is desired to increase the flexibility of layers without increasing the number of rounding operations. We introduce a \textit{multivariate} coupling transformation called Lower Triangular Coupling, which is specifically designed such that the number of rounding operations remains unchanged. In practice, Lower Triangular Coupling does not offer significant improvements over standard coupling layers, and they both attain 4.15 bits per dimension (standard $\pm$0.009 and lower triangular $\pm$0.007), which is averaged over two runs with random weight initialization. The method is presented below for completeness.

The transformation of $\xvec_b$ is formed by multiplication with a strictly lower triangular matrix $\mathbf{L}$ which is conditioned on $\xvec_a$:
% The transformation includes a strictly lower triangular matrix $\mathbf{L}$ which is conditioned on $\xvec_a$, that is multiplied by $\xvec_b$:
\begin{equation}
    \zvec_b = \xvec_b + \left\lfloor \tvec(\xvec_a) + \mathbf{L}(\xvec_a) \xvec_b \right\rceil.
\end{equation}
The main trick is to round the sum of all transformations, such that no additional gradient bias is introduced. This transformation is guaranteed to be invertible, and the inverse can be found with a modified version of forward substitution:
\begin{equation}
    x_i^{(b)} = z^{(b)}_i - \left\lfloor t_i + \sum_{j=1}^{i-1}L_{ij} \cdot x_j^{(b)} \right\rceil,
\end{equation}
where $x_i^{(b)}$ denotes the $i$th element of $\xvec_b$, and $\tvec$ and $\mathbf{L}$ are still conditioned on $\xvec_a$, however, this notation is dropped for clarity. The continuous case can even be solved analytically by using the inverse $\xvec_b = (\mathbf{I} + \mathbf{L})^{-1}\left( \zvec_b - \tvec \right)$.

In practice we restrict the computational cost on feature maps $\xvec, \zvec \in \mathbb{Z}^{n_c \times h \times w}$ by parametrizing a \textit{local} triangular matrix. That is, the transformation can be computed in parallel spatially, and is defined as: $\zvec_{:,vu}^{(b)} = \xvec_{:,vu}^{(b)} + \left\lfloor \tvec_{:,vu} + \mathbf{L}_{vu}\xvec_{:,vu}^{(b)} \right\rceil \forall vu$, where $v,u$ denote spatial coordinates, $\mathbf{L}_{vu} \in \mathbb{R}^{c_b \times c_b}$ and $\tvec$ are conditioned on $\xvec^{(a)}$, and $c_b$ denotes the number of channels in $\xvec^{(b)}$. Since the dimensions of $\mathbf{L}_{vu}$ are small, relative to the neural networks parametrizing them, the inverse can be found in $c_b$ iterations using spatially parallelized matrix operations.

\section{Quantizing a Continuous Flow}
\label{appendix:quantizing_nf}
\begin{wrapfigure}[20]{r}{0.45\textwidth}
    \centering
    \vspace{-5mm}
    \includegraphics[width=0.45\textwidth]{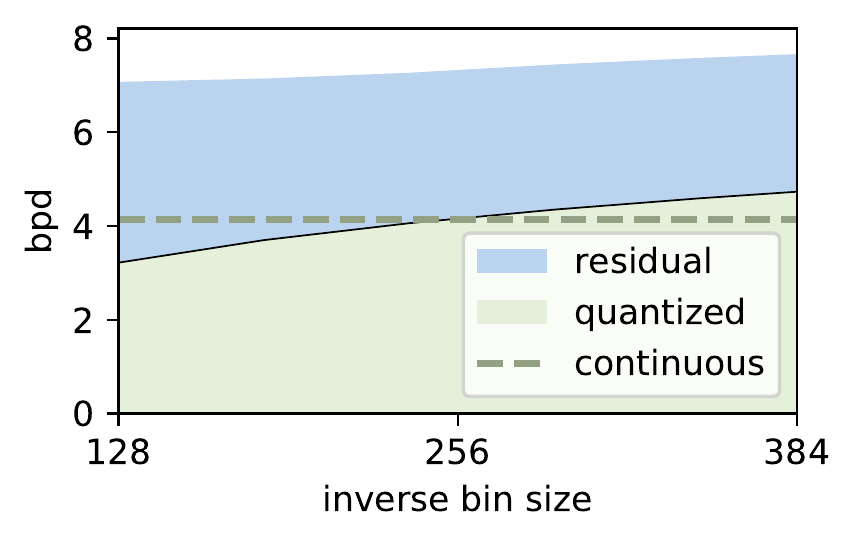}
    \caption{Compression performance of a quantized continuous flow model using different bin sizes. The dashed line denotes the analytical bpd of the continuous model. The total required bpd consists of both the quantized latent $\zvec$ and the residual errors are encoded separately using the FLIF format.}
    \label{fig:quantized_plot}
\end{wrapfigure}

To test the lossless compression performance of continuous flows, the latent space is quantized to a linear spaced bins. Because the latent space is quantized, the reconstructions may contain errors. To enable lossless compression, FLIF is used to encode the errors in reconstruction. Hence, given the quantized latent variables and the reconstruction errors, the original input can be obtained. 

The performance of the quantized flow is shown in Figure \ref{fig:quantized_plot}. When the bin size is large ($\frac{1}{128}$), encoding the latent representation requires relatively few bits, because the probability area is larger. However, the residuals are higher, and require more bits to be modelled. Analogously, when the bin size is small ($\frac{1}{512}$), encoding the latent representation requires more bits, but the residual can be modelled using fewer bits. Although the bits required for the residual or the quantized latents may be small individually, their sum is always large. In total the quantized flow performs poorly on lossless compression.

\section{Experimental details}

\subsection{Networks}
\label{sec:networks}
The coupling and factor out layers are parametrized using neural networks. These networks are DenseNets \cite{huang2017densely}. Specifically we use $n=512$ intermediate channels and a depth $d=12$. In contrast with standard DenseNets, we do not use normalization layers. A single layer in the densenet consists of:
\[
\text{Conv1$\times$1} \to \text{ReLU} \to \text{Conv3$\times$3} \to \text{ReLU},
\]

\subsection{IDF architecture}
\label{sec:idf_architecture}
The exact architecture for experiments is specified in Table \ref{tab:exact_architecture}. All models are trained using Adamax \cite{kingma2014adam} with standard parameters. Furthermore, the learning rate is computed as: $\text{\textit{lr}} = \text{\textit{lr}}_{\text{\textit{base}}} \cdot \text{\textit{decay}}^\text{\textit{epoch}}$. We follow the preprocessing procedure for CIFAR10 as described in \cite{kingma2018glow}. For ImageNet32 and ImageNet64, we do use additional preprocessing. For the ER + BCa dataset, we employ random horizontal and vertical flips during training.

\begin{table}[H]
    \centering
    \caption{IDF architecture and optimization parameters for each experiment.}
    \scalebox{0.74}{
    \begin{tabular}{l r r r r r r r l r }
    \toprule
        Dataset & $L$ & $D$ & densenet depth & densenet channels & batchsize & patchsize & train examples & lr decay & epochs\\ \midrule 
        CIFAR10 & 3 & 8 & 12 & 512 & 256 & 32 & 40000 & 0.999 & 2000 \\
        ImageNet32 & 3 & 8 & 12 & 512 & 256 & 32 & 1230000 & 0.99 & 100 \\
        ImageNet64 & 4 & 8 & 12 & 512 & 64 & 64 & 1230000 & 0.99 & 20 \\
        ER + BCa & 4 & 8 & 12 & 512 & 50 & 80 & 114 & 0.99999 & 50000 \\
        \bottomrule
    \end{tabular}}
    \label{tab:exact_architecture}
\end{table}

In our implementation, instead of using integers in $\mathbb{Z}$, we use the equivalent representation $\mathbb{Z} / 256$, which we found to work better with standard weight initialization and optimization methods. Despite the fact that this implementation does not use integers, it is functionally equivalent to the method presented in the main text.

\subsection{Dataset preparation}
\label{sec:dataset_preparation}
The dataset for CIFAR10 originally consists of 50000 train images and 10000 test images. We use the last 10000 images for validation which results in 40000 train, 10000 validation and 10000 test images. ImageNet32 and ImageNet64 originally contain approximately 1250000 train and 50000 validation images. The validation images are used solely for testing, and 20000 images are randomly selected as a new validation set. This results in roughly 1230000 train, 20000 validation and 50000 test images.

The ER + BCa dataset \cite{janowczyk2018resolution} \footnote{http://andrewjanowczyk.com/wp-static/nuclei.tgz} is split into $114$ train images and $28$ test images such that specific patients IDs only occur in one of the two sets. The test patient identifiers are: 
%\texttt{[8915, 8959, 9023, 9081, 9256, 9382, 10264, 10301, 12749, 16532, 12818, 12871, 12884, 12908, 12931, 12949, 13106, 13459, 13459, 13617, 13694, 14154, 14305, 16661, 17117, 17643, 25289, 25617]}.
\begin{verbatim}
              8915  8959  9023  9081  9256  9382  10264 10301
              12749 16532 12818 12871 12884 12908 12931 12949
              13106 13459 13459 13617 13694 14154 14305 16661
              17117 17643 25289 25617
\end{verbatim}

\subsection{Hardware and Software}
The code for our experiments is implemented using PyTorch~\cite{paszke2017automatic}.
The model implementations are based on the codebase released along with~\cite{van2018sylvester} whereas the rANS coder implementation was taken from~\cite{townsend2019practical}. All experiments were run using 4 Nvidia GTX 1080Ti GPUs.
\label{appendix:experimental_details}

\end{document}